\documentclass[letterpaper]{article} 

\usepackage{main}
\usepackage{times}
\usepackage{helvet}
\usepackage{courier}
\usepackage[hyphens]{url}
\usepackage{graphicx}

\usepackage{natbib}
\usepackage{caption}

\usepackage{amsmath}
\usepackage{amssymb}
\usepackage{multirow}
\usepackage{booktabs}
\usepackage{xspace}
\usepackage[table]{xcolor}
\newcommand{\eg}{\textit{e.g.}\xspace}

\usepackage{algorithm}
\usepackage{algorithmic}

\usepackage{newfloat}
\usepackage{listings}
\DeclareCaptionStyle{ruled}{labelfont=normalfont,labelsep=colon,strut=off}
\floatstyle{ruled}
\newfloat{listing}{tb}{lst}{}
\floatname{listing}{Listing}

\usepackage{xcolor}

\definecolor{link}{RGB}{255,128,134}
\definecolor{la}{RGB}{0,158,57}
\usepackage[pagebackref=false,breaklinks,colorlinks,allcolors=magenta]{hyperref}

\usepackage{cleveref}

\title{\textit{Veila}: Panoramic LiDAR Generation from a Monocular RGB Image}

\author{
Youquan Liu\textsuperscript{\rm 1}\quad 
Lingdong Kong\textsuperscript{\rm 2}\quad
Weidong Yang\textsuperscript{\rm 1,}\thanks{Corresponding authors.}\quad
Ao Liang\textsuperscript{\rm 2}\quad
Jianxiong Gao\textsuperscript{\rm 1}\\
Yang Wu\textsuperscript{\rm 3}\quad
Xiang Xu\textsuperscript{\rm 4}\quad
Xin Li\textsuperscript{\rm 5}\quad
Linfeng Li\textsuperscript{\rm 2}\quad
Runnan Chen\textsuperscript{\rm 6,}\footnotemark[1]\quad
Ben Fei\textsuperscript{\rm 7,}\footnotemark[1]
}
\affiliations{
\textsuperscript{\rm 1}Fudan University \quad
\textsuperscript{\rm 2}National University of Singapore \quad
\textsuperscript{\rm 3}Nanjing University of Science and Technology \\
\textsuperscript{\rm 4}Nanjing University of Aeronautics and Astronautics \quad
\textsuperscript{\rm 5}Shanghai AI Laboratory \\ 
\textsuperscript{\rm 6}University of Sydney \quad
\textsuperscript{\rm 7}The Chinese University of Hong Kong
}

\begin{document}
\maketitle

\begin{figure*}[t]
    \centering
    \includegraphics[width=\textwidth]{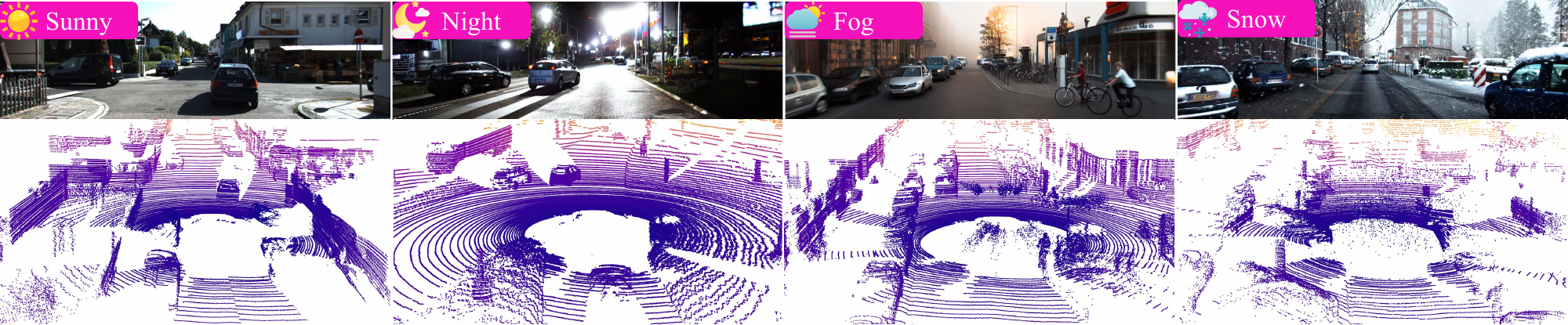}
    \caption{Motivation for \textit{\textbf{Veila}} for panoramic LiDAR generation from monocular images. The top row shows monocular RGB images used as conditions under various weather settings (\eg, sunny, night, fog, and snow). The bottom row presents panoramic LiDAR scans generated by our framework, enabling controllable high-fidelity cross-weather synthesis.}
\label{fig:teaser}
\end{figure*}

\begin{abstract}
\label{sec:abs}
Realistic and controllable panoramic LiDAR data generation is critical for scalable 3D perception in autonomous driving and robotics. Existing methods either perform unconditional generation with poor controllability or adopt text-guided synthesis, which lacks fine-grained spatial control. Leveraging a monocular RGB image as a spatial control signal offers a scalable and low-cost alternative, which remains an open problem. However, it faces three core challenges: (i) semantic and depth cues from RGB are vary spatially, complicating reliable conditioning generation; (ii) modality gaps between RGB appearance and LiDAR geometry amplify alignment errors under noisy diffusion; and (iii) maintaining structural coherence between monocular RGB and panoramic LiDAR is challenging, particularly in non-overlap regions between images and LiDAR. To address these challenges, we propose \textit{\textbf{Veila}}, a novel conditional diffusion framework that integrates: a Confidence-Aware Conditioning Mechanism (CACM) that strengthens RGB conditioning by adaptively balancing semantic and depth cues according to their local reliability; a Geometric Cross-Modal Alignment (GCMA) for robust RGB-LiDAR alignment under noisy diffusion; and a Panoramic Feature Coherence (PFC) for enforcing global structural consistency across monocular RGB and panoramic LiDAR. Additionally, we introduce two metrics, Cross-Modal Semantic Consistency and Cross-Modal Depth Consistency, to evaluate alignment quality across modalities. Experiments on nuScenes, SemanticKITTI, and our proposed KITTI-Weather benchmark demonstrate that Veila achieves state-of-the-art generation fidelity and cross-modal consistency, while enabling generative data augmentation that improves downstream LiDAR semantic segmentation.
\end{abstract}
\section{Introduction}
\label{sec:intro}

LiDAR point clouds are indispensable for 3D perception in autonomous driving and robotics, facilitating critical tasks such as 3D scene understanding~\cite{chen2023clip2Scene,liu2023seal,kong2023laserMix,xu20244d}. However, acquiring large-scale, high-quality LiDAR data in diverse environments is prohibitively expensive and time-consuming~\cite{wu2025weathergen,sun2020waymo}. This motivates the development of generative models to synthesize diverse, high-fidelity, and controllable LiDAR data as a cost-effective alternative.

Among image generative models~\cite{cao2024survey,wang2021generative}, diffusion models~\cite{saharia2022photorealistic} have emerged as a powerful paradigm for high-quality data synthesis due to their ability to model complex distributions and support fine-grained controllability \cite{zhang2023text,jia2024ssmg,zhan2024conditional}, underscoring their potential as a paradigm for generating LiDAR data characterized by sparsity and irregular structures~\cite{zhu2021cylinder3d,kong2025lasermix++,bian2025dynamiccity,kong2023rethinking,liang2025pi3det,hong2024dsnet4d}. 

Recent explorations have extended diffusion models to LiDAR data generation, demonstrating promising initial results. For instance, LiDARGen~\cite{zyrianov2022lidargen} and RangeLDM~\cite{hu2024rangeldm} utilize diffusion models in range-view space but provide limited controllability, while Text2LiDAR~\cite{wu2024text2lidar} employs text-driven conditioning but lacks fine-grained spatial control. Despite recent progress, more flexible and fine-grained conditional LiDAR generations are required.

Image-conditioned LiDAR generation might be a promising alternative. Monocular cameras are widely deployed in cost-sensitive and modern autonomous platforms~\cite{chen2016monocular}, enabling scalable and low-cost data collection. The captured RGB images provide the necessary semantic and depth-related cues, which are complementary to each other. Semantic information encodes object categories and scene layouts, facilitating scene-level understanding~\cite{chen2017rethinking}. 
Meanwhile, monocular RGB images implicitly contain depth-related visual cues (\textit{e.g.}, perspective, occlusion, and scale) that serve as structural priors and depth ordering for 3D reconstruction~\cite{yang2024depth}.

This raises a critical research question: \textit{\textbf{How do diffusion models generate a high-fidelity panoramic LiDAR scene from a monocular RGB image?}}

Despite monocular RGB images providing a potential alternative to generate panoramic LiDAR, it has three key challenges: (1) How to combine both semantic and depth cues to construct a reliable conditioning signal? Semantic and depth cues extracted from the image exhibit complementary strengths but spatially varying reliability, i.e., semantic cues perform better in textured regions, while depth cues are more stable in geometrically structured or textureless areas. Relying on either cue alone is suboptimal; (2) Maintaining robust cross-modal alignment between RGB appearance and LiDAR geometry is non-trivial, due to their inherent modality gap and the progressive noise in intermediate diffusion stages, which often leads to correspondence collapse and structural distortion; (3) Ensuring structural consistency across a LiDAR panorama is particularly challenging in regions beyond the RGB field of view, where the absence of conditioning signals can lead to geometric drift or discontinuities.

To address these gaps, we propose \textit{\textbf{Veila}}, a novel conditional diffusion framework for generating high-fidelity panoramic LiDAR scenes from a monocular RGB image. It introduces three key components:
\\
\noindent\textbf{(1)} A Confidence-Aware Conditioning Mechanism (CACM) adaptively integrates the strengths of semantic and depth information from RGB images to obtain a reliable conditioning signal; 
\\
\noindent\textbf{(2)} A Geometric Cross-Modal Alignment (GCMA) module that leverages epipolar geometry~\cite{hartley2003multiple,tseng2023consistent} to maintain robust RGB-LiDAR alignment throughout noisy diffusion stages; 
\\
\noindent\textbf{(3)} A Panoramic Feature Coherence (PFC) strategy that enforces structural consistency across the generated panoramic LiDAR. By simultaneously ensuring reliable conditioning, robust cross‑modal alignment, and global panoramic coherence, Veila addresses the fundamental challenges of RGB‑to‑LiDAR generation, thereby producing panoramic LiDAR scenes that are faithful to the RGB input and structurally consistent across the full field of view.

Moreover, since existing evaluation protocols lack metrics to assess the consistency between generated LiDAR and monocular RGB conditions, we propose two novel metrics called Cross-Modal Semantic Consistency (CM-SC) and Cross-Modal Depth Consistency (CM-DC), which quantitatively evaluate cross-modal alignment.

Additionally, due to the current datasets containing limited RGB-LiDAR pairs under adverse weather conditions, the applicability to generate real-world scenarios remains unverified. Therefore, to explore the ability of generating adverse weather LiDAR scenes as illustrated in \Cref{fig:teaser}, we present the KITTI-Weather dataset, which contains scenes in clean, foggy, snowy, and nighttime settings. Extensive experiments on nuScenes, SemanticKITTI, and KITTI-Weather benchmark demonstrate that our framework achieves state-of-the-art performance in both fidelity and cross-modal consistency.  Moreover, it significantly improves downstream LiDAR segmentation.

\noindent The main contributions of this paper are summarized:
\begin{itemize}
    \item We propose \textit{\textbf{Veila}}, the first diffusion framework for generating panoramic LiDAR under the guidance of monocular RGB images, addressing key challenges including reliable control signal extraction, noisy cross-modal alignment, and global structural coherence.
    
    \item We design three novel modules: \textit{i)} CACM for acquiring complementary semantic and depth conditioning signals; \textit{ii)} GCMA module for maintaining robust RGB-LiDAR alignment under noisy diffusion; and \textit{iii)} PFC strategy to enforce global spatial consistency.
    
    \item We introduce a modality consistency evaluation protocol comprising two novel metrics, CM-SC and CM-DC, along with the KITTI-Weather benchmark to assess LiDAR generation under adverse weather conditions.
    
    \item We demonstrate the effectiveness of our method through extensive experiments on SemanticKITTI, nuScenes, and KITTI-Weather, achieving the state-of-the-art fidelity and improving downstream LiDAR semantic segmentation by generative data augmentation.  
\end{itemize}
\section{Related Work}
\label{sec:related_work}

\noindent\textbf{Denoising Diffusion Models.}
Diffusion models (DDMs) have emerged as a powerful generative paradigm for generative modeling, achieving state-of-the-art results across 2D vision and 3D domains. Early works~\cite{song2020denoising, ddpm} performed denoising directly in pixel space for high-quality image synthesis. To improve efficiency, Latent Diffusion Models (LDMs)~\cite{latentdiffusion} operate in perceptually compressed latent spaces. Transformer-based DDMs (\eg, DiT~\cite{peebles2023scalable}) scale diffusion to higher resolutions with enhanced modeling capacity. Recent efforts extend DDMs to controllable generation~\cite{zhang2023adding} and 3D-aware tasks for 3D point cloud synthesis~\cite{luo2021diffusion, zhou20213d}, demonstrating their versatility for complex modalities like LiDAR.

\noindent\textbf{LiDAR Scene Generation.}
Generative models for LiDAR data remain relatively underexplored. LiDARGen~\cite{zyrianov2022lidargen} pioneered diffusion-based LiDAR generation by learning the score function in range-view space. Building on this, LiDM~\cite{ran2024lidm} improved LiDAR geometric using latent diffusion with structural preservation. R2DM~\cite{nakashima2024r2dm} refined diffusion architectures and analyzed fidelity-critical components for unconditional scene generation, while UltraLiDAR~\cite{xiong2023ultralidar} adopted a VQ-VAE~\cite{van2017neural} framework for LiDAR completion. RangeLDM~\cite{hu2024rangeldm} targeted real-time efficiency, and Text2LiDAR~\cite{wu2024text2lidar} introduced text-driven control for scene synthesis. However, prior works primarily address unconditional generation or text conditioning, monocular RGB-conditioned panoramic LiDAR generation remains largely unexplored.

\noindent\textbf{Multi-Modal Feature Fusion.}
RGB-LiDAR fusion is pivotal for 3D scene understanding, integrating the semantic richness of RGB images with the geometric accuracy of LiDAR data. Projection-based methods~\cite{wang2021pointaugmenting, krispel2020fuseseg, zhuang2021perception} map features across modalities using calibration transformations to enable spatial alignment. Attention-based approaches~\cite{li2022deepfusion, li2023logonet, liu2023uniseg} utilize LiDAR features as queries to adaptively attend to RGB information for fine-grained integration. Recently, Mamba-based models~\cite{wang2025mambafusion, gu2023mamba} employed linear state-space models (SSMs) for efficient long-range dependency modeling and scalable cross-modal fusion. Nonetheless, these methods depend on precise alignment and are unsuitable for diffusion-based generation, where intermediate representations often exhibit severe noise.
\section{Methodology}
\label{sec:method}

\begin{figure*}[t]
    \centering
    \includegraphics[width=\textwidth]{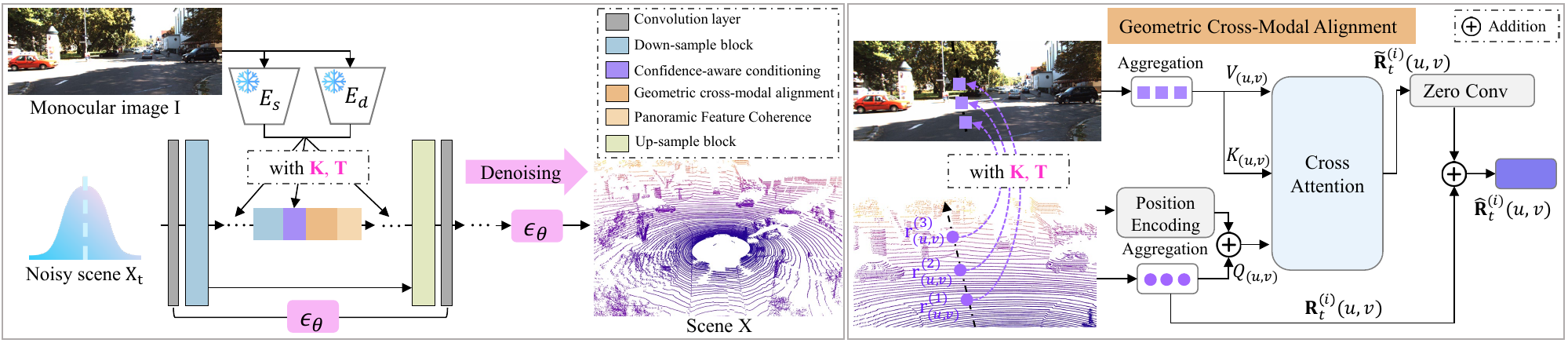}
    \vspace{-0.5cm}
    \caption{\textbf{Method overview.} (\textit{left}) Given a noisy scene $\text{X}_t$ at diffusion step $t$, our framework progressively denoises it into a panoramic LiDAR scene X, conditioned on a monocular RGB image I. Three key designs are embedded within the U-Net backbone of the diffusion model: i) Confidence-Aware Conditioning Mechanism (CACM) adaptively integrates semantic and depth features from frozen encoders $E_s$ and $E_d$; ii) Geometric Cross-Modal Alignment (GCMA) ensures robust alignment between RGB and LiDAR domains using the known camera matrices $\mathbf{K}$ and $\mathbf{T}$; and iii) Panoramic Feature Coherence (PFC) enforces global structural consistency across the panorama. (\textit{right}) Illustration of GCMA: we show the LiDAR ray corresponding to a range image coordinate $(u,v)$ as an example, highlighting how GCMA leverages epipolar geometry to facilitate cross-modal alignment under noisy diffusion stages.}
    \label{fig:framework}
\end{figure*}

In this section, we present \textit{\textbf{Veila}}, a conditional diffusion framework for panoramic LiDAR generation from a monocular RGB image as shown in \Cref{fig:framework}. Our \textit{\textbf{Veila}} comprises three core components: a Confidence-Aware Conditioning Mechanism, a Geometric Cross-modal Alignment module, and a Panoramic Feature Coherence strategy, addressing key challenges in RGB-conditioned LiDAR generation. The following subsections describe each component in detail.

\subsection{Preliminaries}
\noindent\textbf{Range Image Representation.}
A LiDAR point cloud is defined as \(\mathcal{P} = \{ (p^m, e^m) \mid m = 1, ..., N \}\), where each point \(p^m \in \mathbb{R}^3\) represents 3D coordinates, and \(e^m \in \mathbb{R}^L\) denotes auxiliary $L$ attributes such as intensity. Following prior works~\cite{nakashima2024r2dm,wu2024text2lidar,xu2025frnet}, we transform \(\mathcal{P}\) into a structured range image \(\text{X} \in \mathbb{R}^{h \times w \times 2}\) via spherical projection~\cite{milioto2019rangenet}:
\begin{equation}
\left( \begin{array}{c} u \\ v \end{array} \right) = 
\left( \begin{array}{c} 
\frac{1}{2} \left[ 1 - \arctan(p_y^m, p_x^m) {\pi}^{-1} \right] w \\
\left[ 1 - (\arcsin(p_z^md^{-1}) + f_{\text{down}})f^{-1} \right] h
\end{array} \right)~,
\label{eq:spherical_projection}
\end{equation}
where $(u,v)$ are range image coordinates, $(h,w)$ are the height and width of the range image, $f=|f_{\text{up}}|+|f_{\text{down}}|$ is the vertical field-of-view of the sensor, and $d = \|p^m\|_2$ is the Euclidean distance. Each pixel encodes depth and intensity values, allowing standard 2D convolutional architectures to process 3D point cloud data effectively.

\noindent\textbf{LiDAR-Camera Projection.}
To establish the alignment between LiDAR point clouds and camera image pixels, we project 3D LiDAR coordinates into 2D image coordinates. Given a 3D point $p^m = (p_x^m, p_y^m, p_z^m)$ in the LiDAR coordinate system, the corresponding RGB image coordinate $(u', v')$ are obtained by:
\begin{equation}
[u', v', 1]^{\top} = \frac{1}{p_z^m} \cdot \mathbf{K} \mathbf{T} \cdot [p_x^m, p_y^m, p_z^m, 1]^{\top},
\label{eq:camera_projection}
\end{equation}
where $\mathbf{T} \in \mathbb{R}^{4 \times 4}$ denotes the camera extrinsic matrix and $\mathbf{K} \in \mathbb{R}^{3 \times 4}$ is the camera intrinsic matrix, and $[\cdot]^{\top}$ denotes the matrix transpose operation.

\subsection{Problem Formulation}
We regard RGB-conditioned panoramic LiDAR generation as a conditional diffusion process. Given a monocular RGB image \(\text{I} \in \mathbb{R}^{H \times W \times 3}\) as condition, our goal is to generate a corresponding panoramic LiDAR point cloud represented as range image \(\text{X} \in \mathbb{R}^{h \times w \times 2}\), where each pixel encodes depth and intensity. Following the denoising diffusion probabilistic model (DDPM)~\cite{ddpm} framework, we learn to predict the Gaussian noise \(\boldsymbol{\epsilon}\) added to clean data \(\text{X}_0\) through a noise prediction network \(\boldsymbol{\epsilon}_\theta\). The training objective minimizes:
\begin{equation}
\mathcal{L} = \mathbb{E}_{\text{X}_0, \boldsymbol{\epsilon} \sim \mathcal{N}(0,1), t}\left[\|\boldsymbol{\epsilon} - \boldsymbol{\epsilon}_\theta(\text{X}_t, \text{I}, t)\|^2\right],
\end{equation}
where \(\text{X}_t\) is the noisy sample at timestep \(t\). This formulation presents key challenges in constructing reliable conditioning signals, maintaining cross-modal alignment, and enforcing global structural coherence, which are addressed by the components introduced in the following sections.

\subsection{Confidence-Aware Conditioning}
RGB images provide both semantic and depth cues, which offer complementary guidance but vary in spatial reliability. Relying on either alone leads to suboptimal conditioning. To this end, we introduce CACM, which adaptively fuses semantic and depth features based on their local reliability.

\noindent\textbf{Semantic and Depth Feature Extraction.} 
Semantic information encodes object categories, scene layout, and contextual relationships, while depth cues reflect depth ordering and structural priors. To extract these complementary signals, we employ two parallel frozen encoders: a semantic encoder $E_s$ and a depth encoder $E_d$, both pretrained on large-scale vision datasets. Freezing these encoders preserves pretrained knowledge and mitigates overfitting on limited LiDAR data. The semantic encoder $E_s$ produces multi-scale feature maps $\{F_s^{(i)}\}_{i=1}^4$, with each $F_s^{(i)} \in \mathbb{R}^{H_i \times W_i \times D_i^s}$ capturing semantic context at scale $i$. Similarly, the depth encoder $E_d$ outputs depth-aware features $\{F_d^{(i)}\}_{i=1}^4$, where $F_d^{(i)} \in \mathbb{R}^{H_i \times W_i \times D_i^d}$ encodes geometric structure.

\noindent\textbf{Adaptive Conditioning from RGB Features.} 
Although semantic and depth features provide complementary information, their spatial reliability varies: semantic cues are more informative in textured and semantically rich regions, while depth estimates are more stable in geometrically structured or textureless areas. This spatial complementarity indicates that neither cue alone is sufficient. Naive fusion strategies such as concatenation fail to capture this heterogeneity, leading to suboptimal conditioning. In diffusion-based generation, such inconsistencies may accumulate across denoising steps, resulting in structural distortions or semantic artifacts. 

To address this, we propose a Confidence-Aware Conditioning Mechanism that adaptively integrates semantic and depth cues based on local reliability. Since semantic features preserve finer spatial detail, we first interpolate depth features to match the semantic resolution and project both to a shared latent space:
\begin{equation}
\tilde{F}_s^{(i)} = \text{Conv}(F_s^{(i)}), \quad \tilde{F}_d^{(i)} = \text{Conv}(\text{Interp}(F_d^{(i)})),
\end{equation}
where $\text{Interp}(\cdot)$ denotes bilinear interpolation. Confidence scores are computed through separate estimators analyzing local feature statistics:
\begin{equation}
c_s^{(i)} = \text{Sigmoid}(\text{Conv}(\tilde{F}_s^{(i)})), \quad c_d^{(i)} = \text{Sigmoid}(\text{Conv}(\tilde{F}_d^{(i)})).
\end{equation}
The conditioning image features combine both cues with normalized confidence weighting:
\begin{equation}
F^{(i)} = \frac{c_s^{(i)} \cdot \tilde{F}_s^{(i)} + c_d^{(i)} \cdot \tilde{F}_d^{(i)}}{c_s^{(i)} + c_d^{(i)} + \delta},
\end{equation}
where $\delta$ ensures numerical stability. This mechanism allows the network to emphasize more reliable cues in each region, resulting in robust and spatially adaptive conditioning for LiDAR generation.

\subsection{Geometric Cross-Modal Alignment}
The key challenge in RGB-conditioned LiDAR generation lies in establishing accurate correspondences between RGB pixels and range image elements, especially under the noise corruption inherent in diffusion. Conventional projection-based methods depend on current 3D point locations, which become unreliable during intermediate denoising steps~\cite{huang2022multi}. To address this, we propose a Geometric Cross-Modal Alignment module that leverages epipolar constraints derived from geometry-defined LiDAR ray directions (determined solely by the range image coordinates), enabling stable and noise-tolerant correspondences throughout the diffusion process. 

Specifically, each range image pixel $(u,v)$ defines a deterministic ray direction from the LiDAR origin, derived from sensor geometry. Then, we compute the corresponding 3D coordinates $\mathbf{r}_{(u,v)}^{(k)}$ by applying $\text{Ray}(u,v)$, the inverse of~\Cref{eq:spherical_projection}, which recovers the unit 3D point determined by the LiDAR inclination and azimuth angles. These unit points are then scaled by depths $\{d_k\}_{k=1}^{\mathcal{K}}$ to obtain the sampled 3D coordinates:
\begin{equation}
\mathbf{r}_{(u,v)}^{(k)} = d_k \cdot \text{Ray}(u,v).
\end{equation}
These sampled 3D points $\mathbf{r}_{(u,v)}^{(k)}$ are projected to RGB image coordinates using~\Cref{eq:camera_projection}. Features are retrieved from the conditioning representations $F^{(i)}$ at the projected locations $(u'_k, v'_k)$ and aggregated using depth-aware weights:
\begin{equation}
V_{(u,v)} = \frac{\sum_{k=1}^{\mathcal{K}} w_k \cdot F^{(i)}(u'_k, v'_k)}{\sum_{k=1}^{\mathcal{K}} w_k},
\end{equation}
where $w_k = \exp(-d_k/\tau) \cdot m_k$ combines exponential depth decay with validity masks $m_k$.
To enrich the current range image features, we incorporate Fourier positional embeddings~\cite{mildenhall2021nerf} of the 3D coordinates $\mathbf{r}_{(u,v)}^{(k)}$. This encoding captures high-frequency spatial variations, enhancing geometric consistency in cross-modal alignment:
\begin{equation}
Q_{(u,v)} = \mathbf{R}_t^{(i)}(u,v) + \text{MLP}(\gamma(\mathbf{r}_{(u,v)}^{(k)})),
\end{equation}
where $\gamma(\cdot)$ denotes the Fourier positional encoding function applied to 3D coordinates. $\mathbf{R}_t^{(i)}(u,v)$ represents the range image feature at timestep $t$ and scale $i$. The cross-attention output is computed as:
\begin{equation}
\mathbf{\tilde{R}}_t^{(i)}(u,v) = \text{Softmax}\left(\frac{Q_{(u,v)}K_{(u,v)}^{\top}}{\sqrt{d_h}}\right)V_{(u,v)},
\end{equation}
where $K_{(u,v)}$ is set to $V_{(u,v)}$ and $d_h$ is the attention head dimension. The output features $\mathbf{\hat{R}}_t^{(i)}(u,v)$ are integrated via zero-initialized convolutions with residual connections, as illustrated in \Cref{fig:framework}. This design preserves diffusion dynamics while progressively injecting RGB-guided alignment signals.

\subsection{Panoramic Feature Coherence} 
Monocular RGB conditioning primarily affects front-view LiDAR regions where image observations are available. In contrast, rear-view regions lie outside the RGB field of view and therefore lack conditioning signals. During diffusion, these unobserved areas are effectively treated as unconditional generations, often resulting in structural discontinuities or semantic drift across the panoramic scene. 
This challenge is amplified by the limited receptive field of standard UNet-based diffusion architectures, which struggle to propagate global context across spatially distant regions~\cite{vaswani2017attention}. Without explicit regularization, the model may produce inconsistent object structures between conditioned and unconditioned views.

To mitigate this, we propose a Panoramic Feature Coherence strategy that introduces a global self-attention layer at the deepest stage of the UNet, where feature maps have the largest receptive field and capture high-level scene semantics. This component facilitates long-range dependencies, allowing semantic and geometric information from RGB-conditioned regions to propagate throughout the panorama:
\begin{equation}
\mathbf{\bar{R}}_t^{(d)} = \text{SelfAttention}(\mathbf{\hat{R}}_t^{(d)}) + \mathbf{\hat{R}}_t^{(d)},
\end{equation}
where $d$ is the deepest UNet layer. This design promotes spatial coherence across the full panoramic field, helping maintain consistent structures even in unobserved regions. Meanwhile, local fidelity in RGB-visible areas is preserved.
\section{Experiments}
\label{sec:experiments}
\subsection{Experimental Settings}

\begin{table}[t]
    \centering
    \caption{Comparison of \textbf{LiDAR Scene Generation} methods on the \textit{SemanticKITTI} dataset. Lower is better for all metrics ($\downarrow$). MMD is reported in $10^{-4}$.}
    \vspace{-0.2cm}
    \resizebox{0.93\linewidth}{!}{
    \begin{tabular}{r|r|c|c|c|c}
    \toprule
    \textbf{Method} & \textbf{Venue} & \textbf{FRD}$\downarrow$ & \textbf{FPD}$\downarrow$  & \textbf{JSD}$\downarrow$ & \textbf{MMD}$\downarrow$
    \\
    \midrule\midrule
    LiDARGen & ECCV'22 & 735.49 & 119.69  & 0.13 & 21.90
    \\
    LiDM & CVPR'24 & - & 496.78 & 0.08 & 9.20
    \\
    R2DM & ICRA'24 & 262.85 & 12.06  & 0.03 & 0.89
    \\
    Text2LiDAR & ECCV'24 & 567.47  & 16.78 & 0.08 & 4.24
    \\
    \midrule
    \textbf{Veila} & \textbf{Ours} & \textbf{220.40}  & \textbf{8.18} & \textbf{0.02} & \textbf{0.72}
    \\
    \bottomrule
    \end{tabular}}
\label{tab:semkitti_gen}
\vspace{-0.1cm}
\end{table}
\begin{table}[t]
    \centering
    \caption{Comparisons of \textbf{LiDAR Scene Generation} methods on the  \textit{nuScenes} dataset. Metrics with ($\downarrow$) indicate lower is better. The MMD metric is in $10^{-4}$.}
    \vspace{-0.2cm}
    \resizebox{0.93\linewidth}{!}{
    \begin{tabular}{r|r|c|c|c|c}
    \toprule
    \textbf{Method} & \textbf{Venue} & \textbf{FRD}$\downarrow$ & \textbf{FPD}$\downarrow$  & \textbf{JSD}$\downarrow$ & \textbf{MMD}$\downarrow$
    \\
    \midrule\midrule
    LiDARGen & ECCV'22 & 549.18 & 22.80  & 0.04 & 0.76
    \\
    LiDM & CVPR'24 & - & 30.77 & 0.07 & 3.86
    \\
    R2DM & ICRA'24 & 253.80 & 14.35  & \textbf{0.03} & \textbf{0.48}
    \\
    Text2LiDAR & ECCV'24 & 953.18  & 147.48 & 0.09 & 12.50
    \\

    \midrule
    \textbf{Veila} & \textbf{Ours} & \textbf{232.37} & \textbf{12.11} & 0.05 & 0.62
    \\
    \bottomrule
    \end{tabular}}
\label{tab:nusc_gen}
\vspace{-0.1cm}
\end{table}
\begin{table}[t]
    \centering
    \caption{
    Comparisons of \textbf{LiDAR Scene Generation} methods under diverse weather conditions in the \textit{KITTI-Weather} dataset. Metrics with ($\downarrow$) indicate lower is better.
    }
    \vspace{-0.2cm}
    \resizebox{\linewidth}{!}{
    \begin{tabular}{c|r|r|c|c|c}
    \toprule
    \textbf{Type} & \textbf{Method} & \textbf{Venue} & \textbf{FPD}$\downarrow$  & \textbf{JSD}$\downarrow$ & \textbf{MMD}$\downarrow$
    \\
    \midrule\midrule
    \multirow{6.5}{*}{\rotatebox{90}{\textbf{Clean/Night}}}

    & LiDARGen & ECCV'22 & 484.48 & 0.14 & 18.89 
    \\
    & LiDM & CVPR'24 & 127.47 & 0.17 &  31.37
    \\
    & Text2LiDAR & ECCV'24 & 23.61 & 0.06 &  37.63
    \\
    & R2DM & ICRA'24 & 20.27 & \textbf{0.05} &  12.46
    \\
    \cmidrule{2-6}
    & \textbf{Veila (clean)} & \textbf{Ours} & \textbf{13.16} & 0.07 & \textbf{6.30}
    \\
    & \textbf{Veila (night)} & \textbf{Ours} & \textbf{16.31} & 0.07 & \textbf{6.89}
    \\
    \midrule
    \multirow{5.5}{*}{\rotatebox{90}{\textbf{Fog}}}
    & LiDARGen & ECCV'22 & 557.63 & 0.17 &  25.31
    \\
    & LiDM & CVPR'24 & 349.75 & 0.16 &  26.30
    \\
    & Text2LiDAR & ECCV'24 &  40.54 & 0.13 & 19.83 
    \\
    & R2DM & ICRA'24 & 30.27 & \textbf{0.10} & \textbf{11.86}
    \\
    \cmidrule{2-6}
    & \textbf{Veila} & \textbf{Ours} & \textbf{17.57} & \textbf{0.10} & 14.83
    \\
    \midrule
    \multirow{5.5}{*}{\rotatebox{90}{\textbf{Snow}}}
    & LiDARGen & ECCV'22 & 484.48 & 0.16 &  27.66
    \\
    & LiDM & CVPR'24 & 176.73 & 0.13 &  16.70
    \\
    & Text2LiDAR & ECCV'24 &  23.69 & 0.05 &  14.32
    \\
    & R2DM & ICRA'24 & 19.37 & 0.10 & 4.10
    \\
    \cmidrule{2-6}
    & \textbf{Veila} & \textbf{Ours} & \textbf{18.46} & \textbf{0.02} & \textbf{3.30}
    \\\bottomrule
    \end{tabular}}
    \label{tab:kitti_weather}
    \vspace{-0.1cm}
\end{table}

\begin{figure*}[t]
    \centering
    \includegraphics[width=\textwidth]{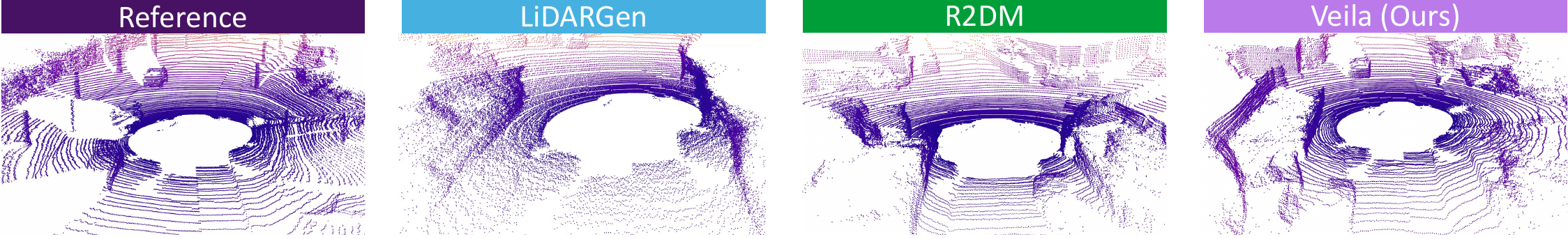}
    \vspace{-0.5cm}
    \caption{Qualitative comparisons of \textit{\textbf{Veila}} against state-of-the-art LiDAR scene generation approaches on the SemanticKITTI dataset. From left to right: Reference (ground truth), LiDARGen, R2DM, and our method.}
    \label{fig:vis_compare}
\end{figure*}

\begin{figure*}[t]
    \centering
    \includegraphics[width=\textwidth]{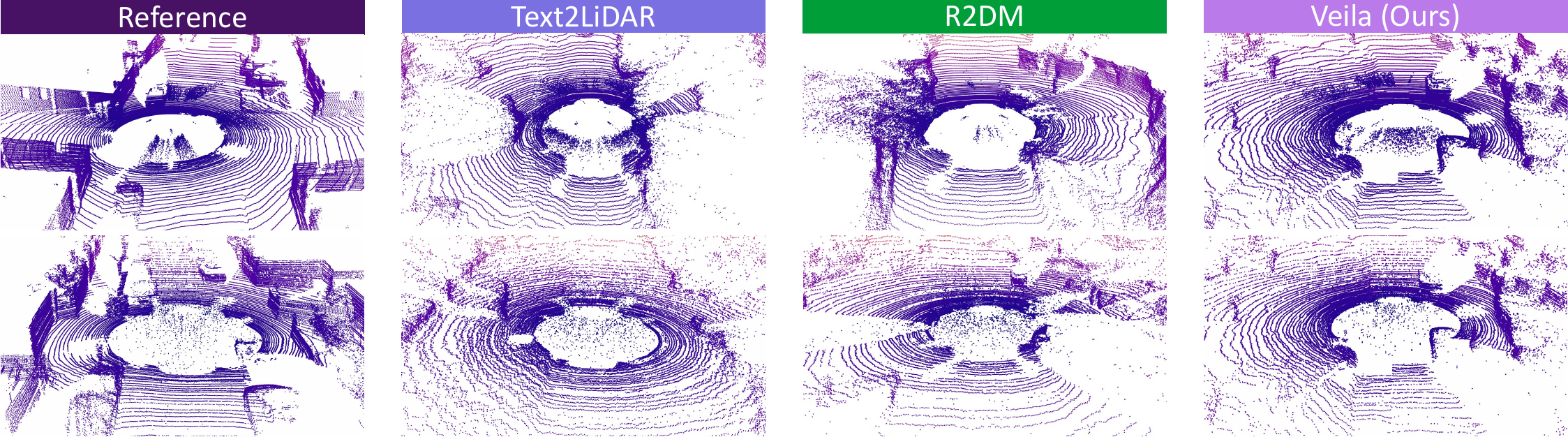}
    \vspace{-0.5cm}
    \caption{Qualitative results on KITTI-Weather. Each row shows LiDAR scenes in different weather conditions: (top) foggy scenes and (bottom) snowy scenes. From left to right: Reference (ground truth), Text2LiDAR, R2DM, and our framework.}
    \label{fig:vis_compare_weather}
\end{figure*}

\begin{table}[!t]
    \centering
    \caption{Comparison of \textbf{LiDAR Scene Generation} methods, with LiDAR scenes partitioned into front-view (\(p_x^m > 0\)) and rear-view (\(p_x^m \leq 0\)) regions on \textit{SemanticKITTI}.}
    \vspace{-0.2cm}
    \resizebox{\linewidth}{!}{
    \begin{tabular}{c|r|ccc|ccc}
    \toprule
    \multirow{2}{*}{\textbf{Method}} & \multirow{2}{*}{\textbf{Venue}} & \multicolumn{3}{c|}{$p_x^m > 0$} & \multicolumn{3}{c}{$p_x^m \leq 0$} \\
    & & \textbf{FPD}$\downarrow$ & \textbf{JSD}$\downarrow$ & \textbf{MMD}$\downarrow$ 
    & \textbf{FPD}$\downarrow$ & \textbf{JSD}$\downarrow$ & \textbf{MMD}$\downarrow$ \\
    \midrule\midrule
    LiDARGen & ECCV'22 & 173.72 & 0.14 & 46.51 & 62.34 & 0.13 & 44.84
    \\
    LiDM & CVPR'24 & 466.44 & 0.08 & 16.94 & 200.05 & 0.09 & 16.93
    \\
    R2DM & ICRA'24 & 11.57 & 0.03 & 0.89 & 12.52 & \textbf{0.04} & 2.37
    \\
    Text2LiDAR & ECCV'24 & 7.87 & 0.07 & 8.29 & 9.61 & 0.08 & 9.44
    \\

    \midrule
    \textbf{Veila} & \textbf{Ours} & \textbf{4.81} & \textbf{0.02} & \textbf{0.74} & \textbf{7.57} & \textbf{0.04} & \textbf{2.19}
     \\
    \bottomrule
    \end{tabular}}
\label{tab:fair}
\vspace{-0.1cm}
\end{table}

\begin{figure*}[t]
    \centering
    \includegraphics[width=\textwidth]{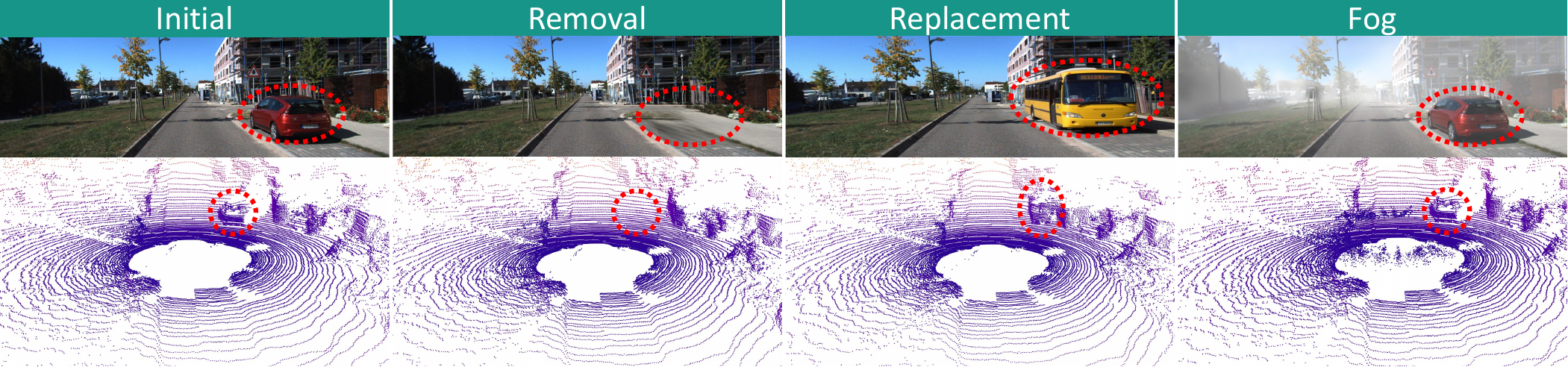}
    \vspace{-0.5cm}
    \caption{Scene editing examples with \textit{\textbf{Veila}}. The highlighted regions mark edited objects and weather modification.}
    \label{fig:vis_control}
\end{figure*}

\begin{figure}[t]
    \centering
    \includegraphics[width=\linewidth]{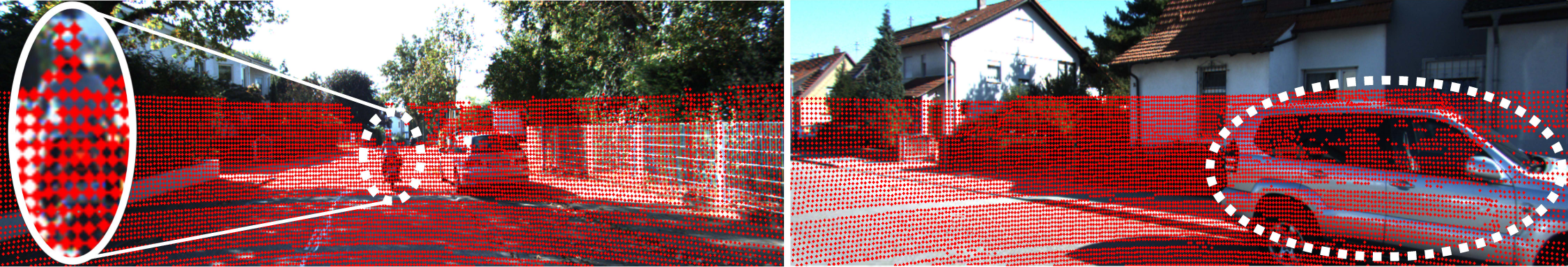}
    \vspace{-0.5cm}
    \caption{LiDAR points generated from \textit{\textbf{Veila}} are projected onto RGB images. Highlighted regions demonstrate precise object-level alignment.}
    \label{fig:vis_projection}
\end{figure}

\noindent\textbf{Datasets.}
We conduct experiments on the nuScenes~\cite{caesar2020nuscenes} and SemanticKITTI~\cite{behley2019semantickitti} datasets. To evaluate generalization under adverse weather, we introduce \textbf{KITTI-Weather}, a new benchmark constructed by augmenting KITTI~\cite{geiger2012kitti} with adverse weather conditions. Specifically, clean RGB-LiDAR pairs from KITTI are used as base scenes, and night, fog, and snow variants are generated using \textbf{DriveGEN}-synthesized RGB images~\cite{lin2025drivegen} and \textbf{Robo3D}-simulated LiDAR point clouds~\cite{kong2023robo3d}. This design enables evaluation of LiDAR generation models in challenging weather scenarios, where such data is scarce.

\noindent\textbf{Evaluation Metrics.}
Following prior work~\cite{nakashima2024r2dm}, we evaluate generation quality using four standard metrics: Fréchet Range Distance (FRD), Fréchet Point Cloud Distance (FPD), Jensen-Shannon Divergence (JSD), and Maximum Mean Discrepancy (MMD). To assess the cross-modal alignment between the generated LiDAR and the RGB input used for conditioning, we introduce two novel metrics: \textbf{Cross-Modal Semantic Consistency (CM-SC)} and \textbf{Cross-Modal Depth Consistency (CM-DC)}. CM-SC measures semantic alignment by projecting per-point semantic labels from the generated LiDAR onto the RGB image plane and comparing them against SegFormer~\cite{xie2021segformer} predictions. CM-DC evaluates depth consistency by projecting LiDAR depths and comparing them with DepthAnything V2 estimates. These metrics evaluate both the structural realism of the generated LiDAR and its semantic and geometric consistency with the RGB input.

\subsection{Quantitative Results}
\noindent\textbf{LiDAR Scene Generation.}
We evaluate our method on SemanticKITTI, nuScenes, and KITTI-Weather, comparing against recent state-of-the-art approaches. Across all benchmarks (\Cref{tab:semkitti_gen}-\ref{tab:kitti_weather}), our method consistently outperforms prior works, achieving lower scores on FRD, FPD, JSD, and MMD. 
On SemanticKITTI (\Cref{tab:semkitti_gen}), our approach reduces FRD by \textbf{16.1\%} and FPD by \textbf{32.1\%} relative to R2DM (Ours: FRD 220.40, FPD 8.18; R2DM: FRD 262.85, FPD 12.06), demonstrating improved range-view fidelity and point-wise accuracy. 
On nuScenes (\Cref{tab:nusc_gen}), it achieves the \textbf{lowest FRD and FPD}, while maintaining competitive JSD, indicating robustness under sparse LiDAR settings. 
In adverse weather scenarios (\Cref{tab:kitti_weather}), our method maintains strong performance across clean, night, fog, and snow conditions. Notably, it achieves a \textbf{$10-25\%$ reduction} in FPD and MMD compared to R2DM and Text2LiDAR, demonstrating its robustness towards adverse weather and domain shifts. These consistent gains highlight the effectiveness of our conditioning and alignment designs in enabling robust, high-fidelity, and structurally coherent LiDAR generation in diverse scenarios.

\noindent\textbf{Region-Wise Generation Quality Assessment.}
To better assess the impact of RGB conditioning, we partition the generated panoramic LiDAR into front-view regions visible to the RGB camera ($p_x^m > 0$) and rear-view regions outside its field of view ($p_x^m \leq 0$). 
As shown in \Cref{tab:fair}, our method achieves superior performance in both regions. While front-view areas benefit directly from RGB guidance, our framework also maintains high fidelity in rear-view regions where no image cues are available. This demonstrates the model’s ability to propagate contextual information beyond the visible region. This capability is enabled by our PFC strategy.

\noindent\textbf{Generative Data Augmentation.} To evaluate the utility of our method for downstream LiDAR segmentation, we generate 10,000 synthetic frames per approach and assign pseudo-labels using a pretrained SPVCNN~\cite{tang2020searching}. These frames are combined with 1\%, 10\%, and 20\% subsets of the SemanticKITTI train set to train two segmentation backbones: MinkUNet~\cite{choy2019minkowski} (voxel-based) and SPVCNN (voxel-point fusion). 
As shown in~\Cref{tab:semkitti_seg}, our method achieves mIoU improvements of $+1.6$ and $+1.3$ over R2DM with MinkUNet and SPVCNN, respectively, in the challenging low-data (1\%) setting. We attribute these gains to the superior fidelity and cross-modal consistency of our generated LiDAR, which provides more informative training signals. These results demonstrate that our model not only produces visually realistic scenes but also yields practical benefits in the downstream perception task.

\subsection{Qualitative Results}
To complement the quantitative results, we present qualitative examples demonstrating our framework’s ability to generate high-fidelity LiDAR scenes, enable controllable scene editing, and maintain cross-modal consistency (\Cref{fig:vis_compare}–\ref{fig:vis_projection}). 

\Cref{fig:vis_compare} shows a qualitative comparison on the SemanticKITTI dataset. LiDARGen produces noisy and scattered point clouds with noticeable structural artifacts, while R2DM yields relatively sparse outputs that lack fine-grained detail. In contrast, our method generates realistic and coherent LiDAR scenes with clearly defined structures (\eg, cars), demonstrating superior fidelity and spatial completeness.
\Cref{fig:vis_compare_weather} showcases results on KITTI-Weather. Compared to Text2LiDAR and R2DM, our framework produces LiDAR scenes that are more structurally consistent with the weather-degraded ground truth, better preserving fine details under foggy and snowy conditions.
\Cref{fig:vis_control} demonstrates our method’s capacity for controllable scene editing, including object removal, replacement, and weather modification. The edited LiDAR scenes maintain structural plausibility and semantic coherence, reflecting the model’s strong geometric understanding across diverse conditions. 
\Cref{fig:vis_projection} depicts the alignment between generated LiDAR and RGB conditions. Highlighted regions show precise object-level alignment, where the generated LiDAR point clouds accurately match object boundaries in the RGB conditions.

\begin{table}[t]
    \centering
    \caption{
        Downstream application of \textit{\textbf{Veila}} to \textbf{LiDAR Semantic Segmentation} task on the \textit{val} set of \textit{SemanticKITTI}. 
    }
    \vspace{-0.2cm}
    \resizebox{0.9\linewidth}{!}{
    \begin{tabular}{c|r|r|ccc}
    \toprule
    \multirow{2}{*}{\textbf{Base}} & \multirow{2}{*}{\textbf{Method}} & \multirow{2}{*}{\textbf{Venue}} & \multicolumn{3}{c}{\textbf{SemanticKITTI}} 
    \\
    & & & \textbf{1\%} & \textbf{10\%} & \textbf{20\%}  
    \\
    \midrule\midrule
    \multirow{5}{*}{\rotatebox{90}{\textbf{MinkUNet}}}
    & \textit{Sup.-only} & - & 40.39 & 60.90 & 62.84 
    \\
    & LiDARGen & ECCV'22 & 36.11 & 54.73 &  60.39
    \\
    & R2DM & ICRA'24 & 53.38 & 60.78 & 62.57
    \\
    & Text2LiDAR & ECCV'24 &  40.23 & 55.00 & 58.35
    \\
    \cmidrule{2-6}
    & \textbf{Veila} & \textbf{Ours} & \textbf{55.01} & \textbf{61.25} & \textbf{63.00}
    \\
    \midrule
    \multirow{5}{*}{\rotatebox{90}{\textbf{SPVCNN}}}
    & \textit{Sup.-only} & - & 37.86 & 59.07 & 61.16 
    \\
    & LiDARGen & ECCV'22 & 36.44 & 55.04 &  59.71
    \\
    & R2DM & ICRA'24 & 50.25 & 60.11 & 62.34
    \\
    & Text2LiDAR & ECCV'24 & 40.55 & 53.87 & 58.34
    \\
    \cmidrule{2-6}
    & \textbf{Veila} & \textbf{Ours} & \textbf{51.53} & \textbf{60.40} & \textbf{62.71} 
    \\\bottomrule
    \end{tabular}}
    \label{tab:semkitti_seg}
    \vspace{-0.1cm}
\end{table}

\begin{table}[!t]
    \centering
    \caption{Ablation study on \textit{SemanticKITTI} evaluating the contribution of each component in \textit{\textbf{Veila}}.}
    \vspace{-0.2cm}
    \resizebox{\linewidth}{!}{
    \begin{tabular}{c|l|c|c|c|c|c|c}
    \toprule
    \textbf{\#} & \textbf{Configuration} & \textbf{FRD}$\downarrow$ & \textbf{FPD}$\downarrow$  & \textbf{JSD}$\downarrow$ & \textbf{MMD}$\downarrow$ &
    \textbf{CM-DC}$\downarrow$ &
    \textbf{CM-SC}$\uparrow$ 
    \\
    \midrule\midrule
    a & $E_s$ only & 301.57 & 21.65 & 0.04 & 1.67 & 0.28 & 60.49
    \\
    b & $E_d$ only & 281.25 & 27.55 & 0.03 & 1.01 & 0.27 & 58.15
    \\
    c & Ours w/o CACM & 342.26 & 17.94 & 0.072 & 4.68 & 0.25 & 58.34
    \\
    d & Ours w/o GCMA & 442.18 &  26.59 & 0.09 & 8.35 & 0.34 & 45.48   \\
    e & Ours w/o PFC & 241.59 & 11.70 & 0.06 & 2.37 & \textbf{0.24} & 63.16
    \\
    \midrule
    \textbf{f} & \textbf{Full Framework} & \textbf{220.40} & \textbf{8.18} & \textbf{0.02} & \textbf{0.72} & \textbf{0.24} & \textbf{63.92} 
    \\\bottomrule
    \end{tabular}}
\label{tab:lidar_ablation}
\vspace{-0.2cm}
\end{table}
\begin{table}[!t]
    \centering
    \caption{Comparison of single-view and multi-view RGB conditioning in \textit{\textbf{Veila}} on the \textit{nuScenes}.}
    \vspace{-0.2cm}
    \resizebox{\linewidth}{!}{
    \begin{tabular}{c|c|c|c|c|c|c|c}
    \toprule
    \textbf{\#} & \textbf{\#Camera(s)} & \textbf{FRD}$\downarrow$ & \textbf{FPD}$\downarrow$  & \textbf{JSD}$\downarrow$ & \textbf{MMD}$\downarrow$ &
    \textbf{CM-DC}$\downarrow$ &
    \textbf{CM-SC}$\uparrow$ 

    \\
    \midrule\midrule
     a & 1 & 232.37 & 12.11 & \textbf{0.05} & 0.62 & 0.34 & 53.77
    \\
    \midrule
     b & 3 & \textbf{215.20} & \textbf{10.40} & \textbf{0.05} & \textbf{0.56} & \textbf{0.29}  & \textbf{58.69}
    \\\bottomrule
    \end{tabular}}
\label{tab:view_ablation}
\vspace{-0.2cm}
\end{table}

\subsection{Ablation Study}
\noindent\textbf{Effects of Each Component.} We evaluate the contribution of each module and feature configuration in \textit{\textbf{Veila}}. As shown in \Cref{tab:lidar_ablation}, using only semantic features from $E_s$ or only depth features from $E_d$ results in consistent degradation across all metrics. This highlights the complementary nature of the two cues: semantic features lack the geometric information required for structural fidelity, while depth features fail to capture semantic context. Removing CACM (w/o CACM) leads to increased FRD and FPD, illustrating the limitations of naive feature concatenation in balancing local semantic and geometric cues. The adaptive weighting in CACM allows the model to dynamically adjust the contributions of semantic and depth features based on their spatial reliability, resulting in more effective conditioning in heterogeneous regions. Excluding GCMA (w/o GCMA) causes a substantial drop in CM-SC and a rise in CM-DC, underscoring its role in maintaining RGB–LiDAR consistency. By enforcing robust geometric alignment between the modalities, GCMA helps mitigate cross-modal misalignment, even under noisy diffusion. 
Finally, removing PFC (w/o PFC) degrades global metrics such as FRD and FPD, indicating that the PFC strategy promotes structural coherence across the panoramic LiDAR, particularly in regions lacking RGB conditioning. Overall, the full framework achieves the best results across all evaluation metrics, confirming that CACM, GCMA, and PFC contribute complementary strengths to enable robust, high-fidelity panoramic LiDAR generation.

\noindent\textbf{Impact of Additional Camera Inputs.} To assess the benefit of multi-view RGB conditioning, we extend \textit{\textbf{Veila}} to incorporate three camera views (front, front-left, and front-right) from \textit{nuScenes}. Each view is independently encoded using $E_s$ and $E_d$, and the resulting features are concatenated along the view dimension before being fed into our method. As shown in \Cref{tab:view_ablation}, incorporating side-view images improves both geometric fidelity and cross-modal consistency compared to using single-view conditioning. We attribute these gains to the expanded field of view and reduced occlusion, which provide richer contextual information for generation.
\section{Conclusion}
\label{sec:conclusion}

We present \textbf{\textit{Veila}}, a novel diffusion framework designed for generating panoramic LiDAR from a monocular RGB image. By systematically addressing key challenges, our method enables realistic and controllable panoramic LiDAR generation. We further propose two cross-modal consistency evaluation metrics and introduce the KITTI-Weather benchmark to standardize assessment under adverse conditions.  
Extensive experiments on SemanticKITTI, nuScenes, and KITTI-Weather datasets show that our method achieves state-of-the-art fidelity and  significantly enhances downstream LiDAR semantic segmentation.

\bibliography{main}

\end{document}